\documentclass[runningheads]{llncs}
\usepackage{graphicx}
\usepackage{subcaption}

\usepackage{tikz}
\usepackage{comment}
\usepackage{amsmath,amssymb} 
\usepackage{color}

\usepackage[accsupp]{axessibility}  
\usepackage{cite}
\usepackage{marvosym}

\begin{document}
\pagestyle{headings}
\mainmatter
\def\ECCVSubNumber{7541}  

\title{AlphaVC: High-Performance and Efficient Learned Video Compression}


\titlerunning{AlphaVC: High-Performance and Efficient Learned Video Compression}
%
\author{Yibo Shi \and
Yunying Ge \and
Jing Wang\textsuperscript{\Letter} \and
Jue Mao}
\authorrunning{Y. Shi et al.}
%
\institute{Huawei Technologies, Beijing China\\
\email{wangjing215@huawei.com}}
\maketitle

\begin{abstract}
  Recently, learned video compression has drawn lots of attention 
  and show a rapid development trend with promising results.
  However, the previous works still suffer from some criticial issues
  and have a performance gap with traditional compression standards 
  in terms of widely used PSNR metric. In this paper, we propose several 
  techniques to effectively improve the performance.
  First, to address the problem of accumulative error, we introduce a 
  conditional-I-frame as the first frame in the GoP, 
  which stabilizes the reconstructed quality and saves the bit-rate.
  Second, to efficiently improve the accuracy of inter prediction without 
  increasing the complexity of decoder, we propose a pixel-to-feature motion prediction method
  at encoder side that helps us to obtain high-quality motion information.
  Third, we propose a probability-based entropy skipping method, 
  which not only brings performance gain, but also greatly reduces the runtime of entropy coding. 
  With these powerful techniques, this paper proposes AlphaVC, a high-performance and efficient learned video compression scheme. 
  To the best of our knowledge, AlphaVC is the first E2E AI codec that exceeds the latest compression standard 
  VVC on all common test datasets for both PSNR (-28.2\% BD-rate saving) and MSSSIM (-52.2\% BD-rate saving), 
  and has very fast encoding (0.001x VVC) and decoding (1.69x VVC) speeds.

\end{abstract}

\section{Introduction}
Video data is reported to occupy more than 82\% of all consumer Internet traffic~\cite{cisco2020cisco},
and is expected to see the rapid rate of growth in the next few years, especially the high-definition videos and ultra high-definition videos.
Therefore, video compression is a key requirement for the bandwidth-limited Internet.
During the past decades, several video coding standards were developed, such as H.264~\cite{wiegand2003overview}, H.265~\cite{sullivan2012overview}, and H.266~\cite{bross2021developments}. 
These methods are based on hand-designed modules such as block partition, inter prediction and transform~\cite{ahmed1974discrete}, etc.
While these traditional video compression methods have made a promising performance, 
their performance are limited since the modules are artificially designed and optimized separately.

Recently, learned image compression~\cite{cui2021asymmetric,cheng2020learned,minnen2018joint,guo2020variable} based on variational auto-encoder~\cite{kingma2013auto} has shown great potential,
achieving better performance than traditional image codecs~\cite{bellard2016bpg,wallace1992jpeg,bross2021developments}. 
Inspired by the learned image compression, and combined with the idea of traditional video codecs, 
many learning-based video compression approaches~\cite{agustsson2020scale,guo2021learning,pourreza2021extending,feng2021versatile,habibian2019video,hu2021fvc,lu2020end,li2021deep} were proposed. 

Given the reference frame, variant kinds of motion compensation (alignment) methods were proposed like 
scale-space alignment~\cite{agustsson2020scale}, feature-based alignment~\cite{hu2021fvc}, multi-scale feature-based alignment~\cite{sheng2021temporal}.
These methods aim to improve the diversity of motion compensation and result in more compression-friendly predictions.
However, such methods increase the complexity on both encoder and decoder side.
Inspired by AMVP (Advanced Motion Vector Prediction) on traditional video compression methods~\cite{sullivan2012overview},
we expect the encoder side to predict a more accurate motion information.
Further, at the encoder side of AlphaVC, we propose a pixel-to-feature motion prediction method that can obtain high-quality 
motion information without increasing the complexity of the decoder.
\begin{figure}
\centering
\includegraphics[width=\linewidth]{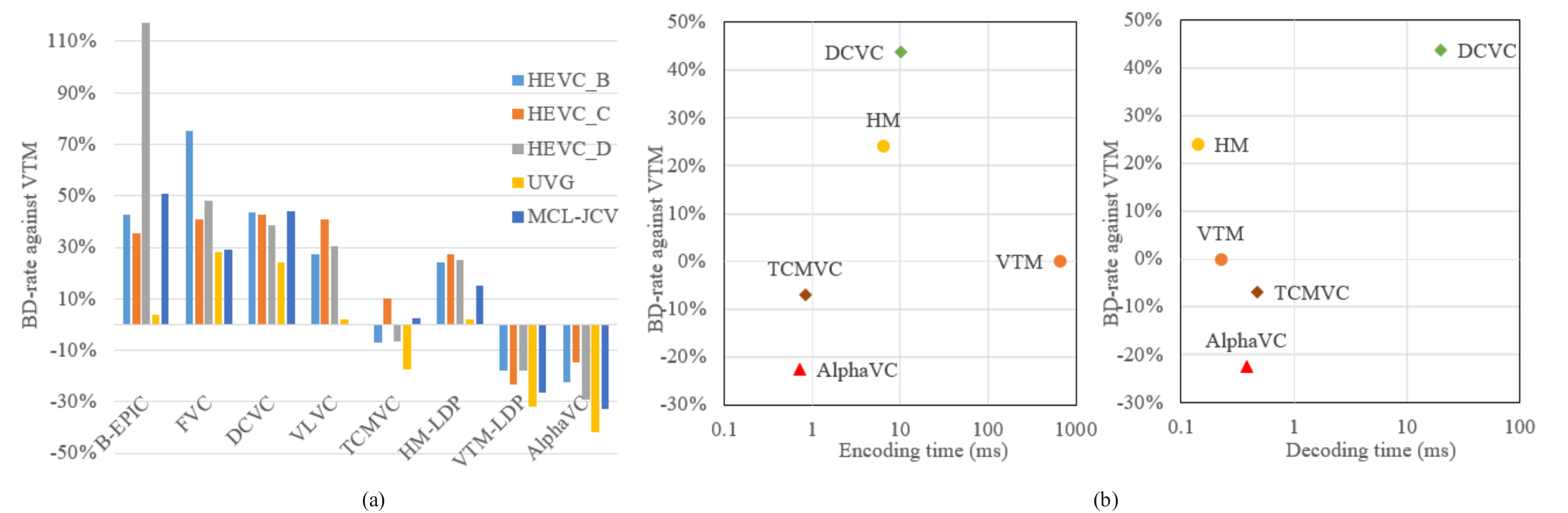}
\caption{
  (a): BD-rate against VTM in terms of PSNR (Lower is better). (b): BD-rate against VTM as a function of encoding/decoding time on 1080p videos.
}
\label{fig:fig1}
\end{figure}

Existing learned video compression can be divided into two categories: Low-Delay P mode and Low-Delay B/Random-Access mode. 
For the Low-Delay P mode, the methods~\cite{agustsson2020scale,guo2021learning,hu2021fvc,sheng2021temporal} only include the P(predictive)-frames and I(image)-frames. 
For the Low-Delay B or Random-Access mode, the methods~\cite{feng2021versatile,pourreza2021extending} insert the B(bidirectional predictive) frames into the GoP to improve compression performance. 
AlphaVC focuses on the Low-Delay P mode. In this mode, due to the accumulation error in P-frame~\cite{lu2020content}, 
most existing methods have to use the inefficient I-frame as the first frame in limited length GoP.
Unlike the existing methods, we overcome this issue by introducing a conditional I-frame (cI-frame) as the first frame in the GoP, 
which stabilizes the reconstructed quality and achieves better performance.

In addition, we all know that the entropy coding~\cite{howard1994arithmetic,duda2009asymmetric} can only run serially will increase the runtime.
Moreover, the auto-regressive entropy module~\cite{minnen2018joint}, which significantly increase the decoding time, 
is always used on learned image codecs for a higher compression ratio.
We found that most elements of the latents usually have very low information entropy, which means the probability distributions of these elements estimated by entropy module always is highly concentrated.
Inspired by this, we propose an efficient probability-based entropy skipping method (Skip) which can significantly save runtime in entropy coding,
and achieve higher performance without auto-regressive.

With the help of the above technologies, AlphaVC achieves the highest E2E compression performance while being very efficient. 
As shown in Fig.~\ref{fig:fig1}, 
the proposed AlphaVC outperforms VTM-IPP/VTM-LDP by 28.2\%/6.59\% , 
where the VTM is the official software of H.266/VVC, 
the IPP denotes the configuration using one reference frame and flat QP,
and the LDP denotes the better configuration using multiple references and dynamic QP.
Note the configuration of AlphaVC is the same as IPP.
To the best of our knowledge, AlphaVC is the only learning-based video codec that can consistently achieve comparable or better performance with VTM-LDP in terms of PSNR on all common test datasets.
Comparing with the state-of-the-art learning-based video codecs~\cite{sheng2021temporal}, AlphaVC reduces the BD-rate by about 25\% while faster encoding and decoding.

Our contributions are summarized as follows:
\begin{enumerate}
  \item We introduce a new type of frame named conditional-I frame (cI-frame) and propose a new coding mode for learned video compression. 
  It can effectively save the bit rate of I-frame and alleviate the problem of accumulated error.
  \item The proposed motion prediction method, utilizing the idea of pixel-to-feature and global-to-local,
  can significantly improve the accuracy of inter-frame prediction without increasing decoding complexity.
  \item An efficient method in entropy estiamtion module and entropy coding have higher performance and faster encoding and decoding time.
\end{enumerate}

\section{Related Work}

\subsection{Image Compression}
In the past decades, the traditional image compression methods like 
JPEG~\cite{wallace1992jpeg}, JPEG2000~\cite{christopoulos2000jpeg2000} and BPG~\cite{bellard2016bpg} can efficiently reduce the image size.
Those methods have achieved a high performance by exploiting the hand-crafted techniques, such as DCT~\cite{ahmed1974discrete}.
Recently, thanks to variational autoencoder~(VAE)~\cite{kingma2013auto} and scalar quantization assumption~\cite{balle2016end}, 
the learning-based image compression methods have achieved great progress.
With the optimization of entropy estimation modules~\cite{balle2018variational,minnen2018joint} and network structure~\cite{cui2021asymmetric,cheng2020learned},
the learning-based image compression methods have achieved better performance than the traditional image compression codecs on common metrics,
such as PSNR and MS-SSIM~\cite{wang2003multiscale}.

\subsection{Video Compression}
Video compression is a more challenging problem compared to image compression.
There is a long history of progress for hand-designed video compression methods, 
and several video coding standards have been proposed, 
such as H.264(JM)~\cite{wiegand2003overview}, H.265(HM)~\cite{sullivan2012overview} and more recently H.266(VTM)~\cite{bross2021developments}. 
With the development of video coding standards, 
the traditional video compression methods made significant improvements and provided a strong baseline.
Even they have shown a good performance, these algorithms are limited by the hand-designed strategy and the difficult to optimize jointly.

Recently, learning-based video compression has become a new direction.
Following the traditional video compression framework, 
Lu et al. proposed the end-to-end optimized video compression framework DVC~\cite{lu2020end}, 
in which the neural networks are used to replace all the critical components in traditional video compression codec.
Then, the exploration direction of existing approaches can be classified into three categories.
One category of approaches focuses on the motion compensation (alignment) method to improve the accuracy of inter prediction.
For example, SSF~\cite{agustsson2020scale} designed a scale-space flow to replace the bilinear warping operation.
Hu et al.~\cite{hu2021fvc} propose the FVC framework, which apply transformation in feature space with deformable convolution~\cite{dai2017deformable}.
Later Sheng et al. introduce multi-scale in feature space transformation~\cite{sheng2021temporal}.
Another popular direction is the design of auto-encoder module.
Such as Habibian et al.~\cite{habibian2019video} use a 3D spatio-temporal autoencoder network to directly compress multiple frames.
Li et al.~\cite{li2021deep} use the predicted frame as the input of encoder, decoder, instead of explicitly computing the residual.
The third category extends the learned video compression to more codec functions, like B-frame~\cite{pourreza2021extending,feng2021versatile}, utilizing multiple reference frames~\cite{hu2021fvc}.

\section{Method}
\subsection{Overview}
Let $\mathcal{X}=\{\mathbf{X}_1, \mathbf{X}_2, \dots\}$ denote a video sequence, 
video codecs usually break the full sequence into groups of pictures (GoP).
Due to the accumulative error of P-frames, in low delay P mode, which is AlphaVC adopted, 
each group needs to start with an I-frame and then follow P-frames.
In AlphaVC, we propose a new codecing mode in GoP, including three types of frames.
As shown in Fig.~\ref{fig:framework}(a), the I-frame is only used for the first frame.
For other groups, we propose to start with conditional-I-frame instead of I-frame.
The Conditional-I-frame (named cI-frame), 
which uses the reference frame as condition of entropy to reduce the bit-rate,
stabilises the reconstructed quality like I-frame, and meanwhile has a high compression rate.
The details of each type of our P-frame and cI-frame are summarized as follows:
\begin{figure}
\centering
\includegraphics[width=1.0\linewidth]{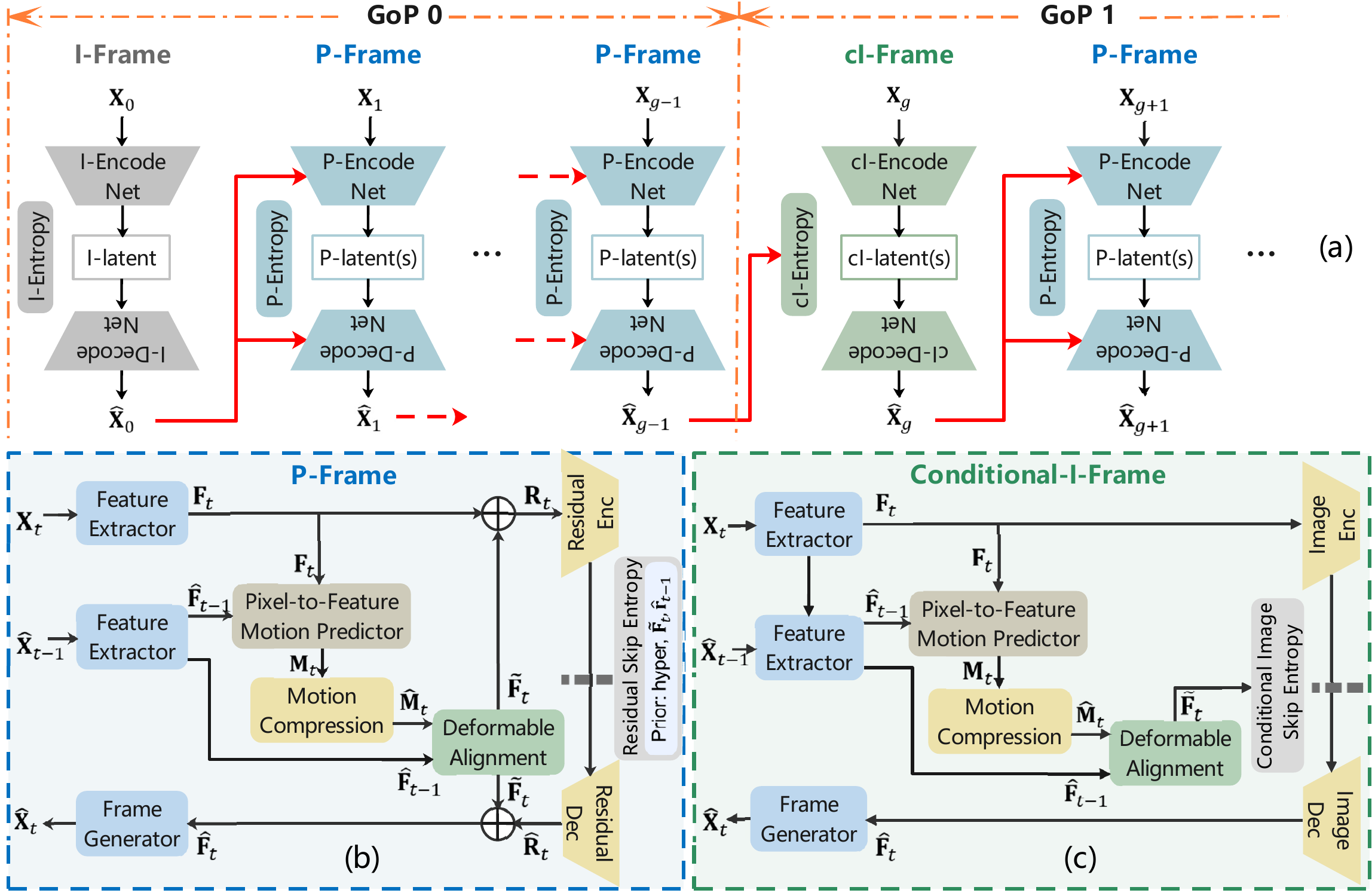}
\caption{Overview of our proposed video compression scheme. (a): Two kinds of GoP. (b): The framework of P-frame. (c): The framework of cI-frame.}
\label{fig:framework}
\end{figure}
\subsubsection{P-Frame} 
First of all, we define the P-Frame in learned video compression as a class of methods that has the following form on decoder side:
\begin{equation} \label{eq:p-frame}
\begin{array}{rl}
  \hat{\bf{X}}_t = D_{p}(H_{\text{align}}(\hat{\bf{X}}_{t-1}, \hat{\bf{m}}_t), \hat{\bf{r}}_t)
\end{array}
\end{equation}
where $D_{p}(\cdot), H_{\text{align}}(\cdot)$ denote the method of reconstruction and alignment, 
$\hat{\bf{m}}_t, \hat{\bf{r}}_t$ are the quantized latent representation of motion, residual.
Note that the quantized latent representation is the features to be encoded after the encoder and quantization.
That is, the reference frame $\hat{\bf{X}}_{t-1}$ will participate in and affect the reconstruction of current frame,
which means that the consecutive P-frame will generate cumulative errors.

In this paper, we use the feature-align based P-frame framework, 
Fig.~\ref{fig:framework}(b) sketches our P-frame compression framework. 
We first transform $\hat{\bf{X}}_{t-1}, \bf{X}_t$ into feature space $\hat{\bf{F}}_{t-1}$, $\bf{F}_t$.
Then motion predictor will generate the predicted motion ${\bf{M}}_t$
and the predicted motion will be compressed by motion compression model.
The predicted feature $\tilde{\bf{F}}_t$ is generated by deformable alignment~\cite{dai2017deformable} 
with the reconstructed motion $\hat{\bf{M}}_t$ and reference feature $\hat{\bf{F}}_{t-1}$.
Finally, the residual in feature-based $\bf{R}_t=\bf{F}_t-\tilde{\bf{F}}_t$ will be compressed by residual compression model.
The reconstructed feature $\hat{\bf{F}}_t=\hat{\bf{R}}_t + \tilde{\bf{F}}_t$ is transformed into the current reconstruct frame $\hat{\bf{X}}_t$ with frame generator.

Both the motion compression model and residual compression model are implemented by auto-encoder structure~\cite{balle2018variational}, 
including an encoder module, decoder module and the proposed entropy estiamtion module.
The newtork structure of auto-encoder part is the same as FVC~\cite{hu2021fvc}. To further reduce redundant information, we introduce the temporal and structure prior 
for the entropy estimation module in both motion and residual compression models:
\begin{equation} \label{eq:entropy-prior}
\begin{array}{rl}
  & \mathbb{E}_{\hat{\bf{m}}_t \sim p_t} [-\log_2{q_t(\hat{\bf{m}}_t | \hat{\bf{F}}_{t-1}, \hat{\bf{m}}_{t-1})}] \\
  & \mathbb{E}_{\hat{\bf{r}}_t \sim p_t} [-\log_2{q_t(\hat{\bf{r}}_t | \tilde{\bf{F}}_{t}, \hat{\bf{r}}_{t-1})}]
\end{array}
\end{equation}

the reference feature $\hat{\bf{F}}_{t-1}$ and previous quantized motion latent representation $\hat{\bf{m}}_{t-1}$ are structure and temporal priors of $\hat{\bf{m}}_t$ respectively, and
the predicted feature $\tilde{\bf{F}}_{t}$ and previous quantized residual latent representation $\hat{\bf{r}}_{t-1}$ are structure and temporal priors of $\hat{\bf{r}}_t$ respectively. 
\subsubsection{Conditional-I-Frame (cI-frame)}
We introduce a new type of frame called the cI-frame like~\cite{liu2020conditional}, which can be formulated as:
\begin{equation} \label{eq:cI-frame}
\begin{array}{rl}
  & \text{Auto-Encoder}: \hat{\bf{y}}_t = Q(E_{cI}({\bf{X}_t})), \hat{\bf{X}}_t = D_{cI}(\hat{\bf{y}}_t), \\
  & \text{Entropy}: R(\hat{\bf{y}}_t|\hat{\bf{X}}_{t-1}) = \mathbb{E}_{\hat{\bf{y}}_t \sim p_t}[-\log_2{q_t(\hat{\bf{y}}_t | H_{\text{align}}(\hat{\bf{X}}_{t-1}, \hat{\bf{m}}_t))}], \\
\end{array}
\end{equation}
where $\hat{\bf{y}}_t$ is the quantized latent representation of $\bf{X}_t$, 
$E_{cI}(\cdot), Q(\cdot), D_{cI}(\cdot)$ denote the function of cI encoder module, quantization and reconstruction. 
That is, cI-frame reduces the inter redundant information through the entropy conditioned on $\hat{\bf{X}}_{t-1}$.
For cI-frame, the input of the autoencoder does not use the reference frames, thus make the reconstructed quality stable.
Further, we use cI-frame as the first frame in the GoP excluding the first GoP, which not only stabilizes the sequence quality like I-frame, 
but also improves the compression ratio, thereby alleviating the problem of accumulated errors.

The framework for cI-frame is shown in Fig.~\ref{fig:framework}(c). 
The feature extractor, motion prediction and motion compression part share the same structure with P-frame framework. 
$\tilde{\bf{F}}_t$ is only used as the prior, the current feature $\bf{F}_t$ will be the only input of the encoder.

Furthermore, we  propose two novel strategies in both P-frame and cI-frame, named pixel-to-feature motion prediction (P2F MP) and probability-based entropy skipping method (Skip),
to improve the accuracy of inter prediction and coding efficiency.
\subsection{Pixel-to-Feature Motion Prediction}
\begin{figure}
\centering
\includegraphics[width=1.0\linewidth]{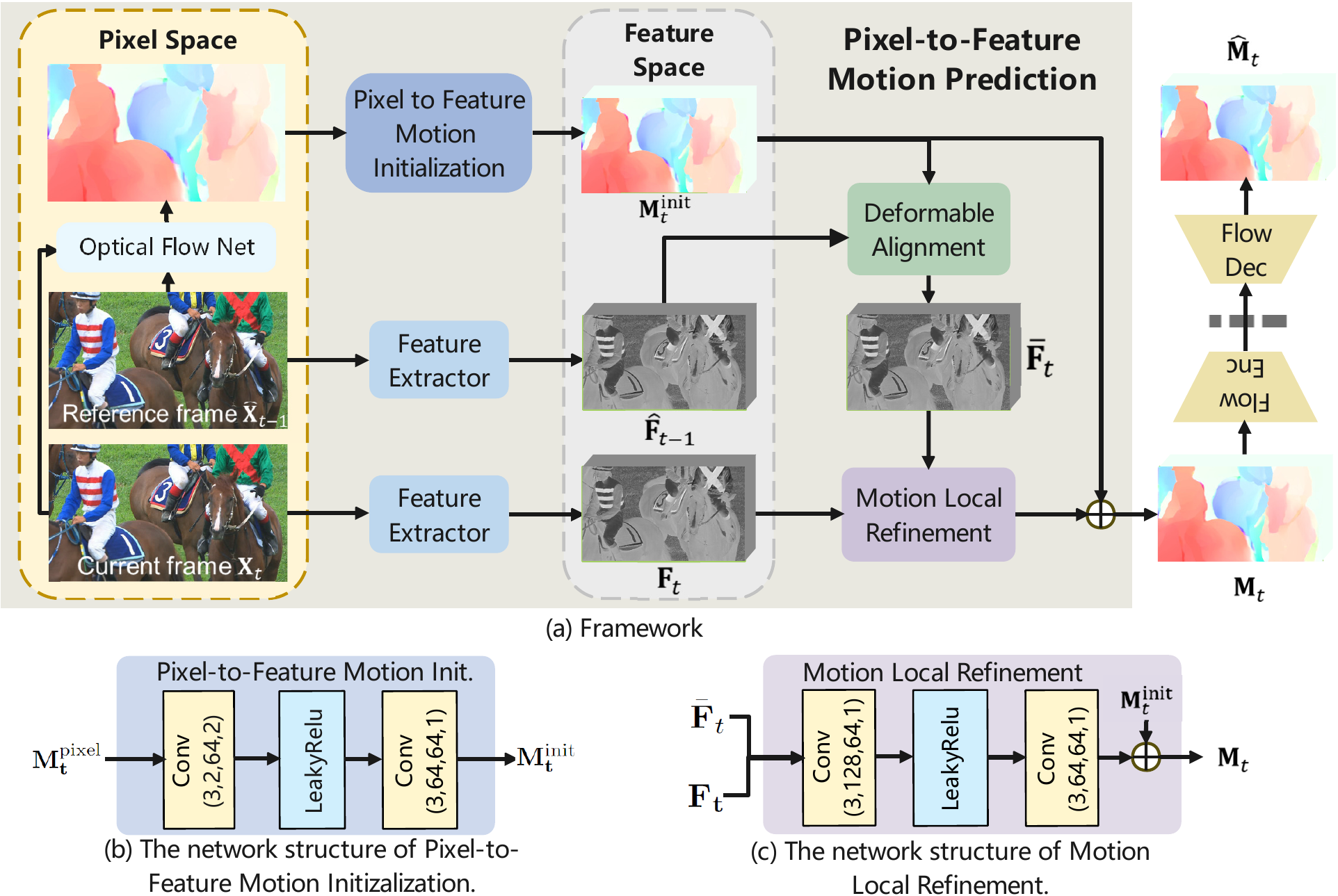}
\caption{Illustration of our proposed pixel-to-feature motion prediction module.}
\label{fig:pixel-to-feature}
\end{figure}
Inter-frame prediction is a critical module to improve the efficiency of inter-frame coding, since it determines the accuracy of the predicted frame.
We propose pixel-to-feature motion prediction to fully exploit 
the diversity of feature-based alignment and the state-of-the-art optical flow network.
The illustration is shown in Fig.~\ref{fig:pixel-to-feature}. 

Given the previous reconstructed frame $\hat{\bf{X}}_{t-1}$ and the current frame ${\bf{X}}_t$,
the optical flow in pixel space $\bf{M}_t^{\text{pixel}}$ will be generated by a state-of-the-art optical flow network~\cite{sun2018pwc,teed2020raft}.
The pixel space motion $\bf{M}_t^\text{pixel}$ is then used to initialize a motion in feature space $\bf{M}_t^{\text{init}}$.
Then, we apply the deformable alignment ${D(\cdot, \cdot)}$ to the reference feature $\hat{\bf{F}}_{t-1}$ by $\bf{M}_t^{\text{init}}$:
\begin{equation} \label{eq:align1}
\begin{array}{rl}
  \bar{\bf{F}}_t = {D}(\hat{\bf{F}}_{t-1}, \bf{M}_t^{\text{init}})
\end{array}
\end{equation}
After initial alignment, the motion local refinement network will refine the initial motion locally 
according to the initially aligned feature $\bar{\bf{F}}_t$ and the target feature $\bf{F}_t$, 
and then generate the final predicted motion $\bf{M}_t$.
\begin{equation} \label{eq:align2}
\begin{array}{rl}
  \bf{M}_t = \text{Refine}(\bar{\bf{F}}_t, F_t) + \bf{M}_t^{\text{init}}
\end{array}
\end{equation}
Finally, the predicted motion will be compressed to reconstruct motion $\hat{\bf{M}}_t$ through motion compression model.

Unlike existing methods, AlphaVC neither learn motion directly from features~\cite{hu2021fvc} that are difficult to fit through convolutions
nor compress the generated optical flow directly~\cite{lu2020end}. We follow pixel-to-feature and global-to-local principles, 
first generate the feature space motion before coding with optical flow, 
then performing further fine-tuning through alignment feedback.
Experiments show that this method greatly improves the accuracy of inter-frame prediction without affecting the decoding complexity and running time. 
\subsection{Probability-base Entropy Skipping Method}

For a latent representation variable $\mathbf{v}$ in learned image or video compression, we first quantize it with round-based quantization $\hat{\mathbf{v}}=[\mathbf{v}]$,
and estimate the probability distribution of ${\mathbf{v}}$ by an entropy estimation module with some priors, such as hyper~\cite{balle2018variational}, context~\cite{minnen2018joint}, etc.
Then $\hat{\mathbf{v}}$ is compressed into the bitstream by entropy coding like arithmetic coding~\cite{howard1994arithmetic}, asymmetric numeral system~\cite{duda2009asymmetric}.
In video compression, due to the introduction of the reference frame, the entropy of quantized latent representation variables like 
$\hat{\mathbf{m}_t}, \hat{\mathbf{r}_t}$ in P-frame is very small, especially in low bit-rate. 
That means the probability distributions of most elements in the latent variable are concentrated.
If it is slightly off-center for such an element, we will encode it to bitstream with a high cost.
In other words, if we skip these elements without encoding/decoding and replace them with the peak of probability distribution, 
we can save both bit-rate and runtime of entropy coding with little error expectations.
Inspired by this idea, we propose an efficient probability-based entropy skipping method (Skip).

For a latent representation variable ${\mathbf{v}}$, 
we define $\mathcal{Q}$ as the probability density set of ${\mathbf{v}}$ estimated by its entropy module.
The value which has the maximum probability density of the $i$-th element is calculated as:
\begin{equation} \label{eq:skip1}
  \theta_{i} = \displaystyle\mathop{\arg \max}_{\theta_{i}} q_i(\theta_{i})
\end{equation}
The probability that the element ${v}_i$ is close to $\theta_i$ can be computed by:
\begin{equation} \label{eq:skip2}
  q_{i}^{\text{max}} = \displaystyle\int_{\theta_{i}-0.5}^{\theta_{i}+0.5} q_i(x) \,dx \\
\end{equation}
If the probability $q_{i}^\text{max}$ is high enough, we will not encode/decode the element to/from the bitstream, and replace the value with $\theta_{i}$. 
After this operation, the quantized latent representation will become $\hat{\mathbf{v}}^\text{s}$:
\begin{equation} \label{eq:skip3}
  \hat{v_{i}}^\text{s} = 
  \left\{ 
    \begin{array}{rl}
      \theta_{i}, & \quad   q_{i}^\text{max} >= \tau \\
      \left[v_i\right], & \quad   q_{i}^\text{max} < \tau
    \end{array}
  \right.
\end{equation}
where $\tau$ is a threshold to determine whether to skip.

In our paper, we use gaussian distribution as the estimated probability density of all the quantized latent representations.
Hence the Eq.~\ref{eq:skip1} and Eq.~\ref{eq:skip2} can be easily solved as:
\begin{equation} \label{eq:skip4}
  \theta_{i} = \mu_i, q_i^\text{max} = \text{erf}(\frac{1}{2\sqrt{2}\sigma_i}).
\end{equation}
It can be seen that $q_i^\text{max}$ is the monotone function of $\sigma_i$, 
we use $\sigma_i$ as the condition of Eq.~\ref{eq:skip3} to further reduce the computational complexity:
\begin{equation} \label{eq:skip5}
  \hat{v_{i}}^\text{s} = 
  \left\{ 
    \begin{array}{rl}
      \mu_{i}, & \quad   \sigma_{i} < {\tau}_{\sigma} \\
      \left[{v}_i\right], & \quad   \sigma_{i} >= {\tau}_{\sigma}
    \end{array}
  \right.
\end{equation}

There are two benefits of Skip. First, it can dynamically reduce the number of elements that need to be entropy encoded, significantly reducing the serial CPU runtime.
Second, we can better trade-off errors and bit rates for elements with high determinism, thereby achieving high compression performance.

\subsection{Loss Function}
Our proposed AlphaVC targets to jointly optimize the rate-distortion (R-D) cost.
\begin{equation} \label{eq:loss-func}
  L=R + \lambda \cdot D = 
  (R_0^\text{I}+\lambda \cdot D_0^\text{I}) + \displaystyle \sum_{t=1}^{\text{T}-1}(R_t^\text{p}+\lambda \cdot D_t^\text{p}) + (R_{\text{T}}^{\text{cI}} + \lambda \cdot D_{\text{T}}^{\text{cI}})
\end{equation}
where the training GoP size is $\text{T}$, $\lambda$ controls the trade-off, 
$R_0^\text{I}-D_0^\text{I}$, $R_t^{\text{p}}-D_t^\text{p}$ and $R_\text{T}^{\text{cI}}-D_\text{T}^\text{cI}$ represent the rate-distortion of the $0$-th I-frame, the $t$-th P-frame and the $\text{T}$-th cI-frame, respectively.

\section{Experiments}
\subsection{Setup}
\subsubsection{Training.}
We train our model on the Vimeo-90k dataset. This dataset consists of 4278 videos with 89800
independent shots that are different from each other in content.
We randomly crop the frames to patches of size $256\times 256$, and start training from scratch.
We train the models with Adam optimizer for 60 epochs,
where the batchsize was set to 8 and learning rate was initially set to $1e-4$ and reduced to half for 30 epochs.
The skip operation will been enabled during training.
The loss function is the joint rate-distortion loss as shown in Eq.~\ref{eq:loss-func}, 
where the multiplier $\lambda$ is chosen from (0.07, 0.05, 0.01, 0.005, 0.001, 0.0007) for the MSE optimization.
The the MS-SSIM optimized models are finetuned from MSE-optimized model with $\lambda=$ 0.03, 0.01, 0.007, 0.005, 0.001.

\subsubsection{Testing.} 
We evaluate our proposed algorithm on the HEVC datasets~\cite{bossen2012common} (Class B,C,D,E), the UVG datasets~\cite{mercat2020uvg}, and the MCL-JCV datasets~\cite{wang2016mcl}.
The HEVC datasets contain 16 videos with different resolution $416\times 240$, $832\times 480$ and $1920\times 1080$.
The UVG and MCL-JVC datasets contain 7 and 30 1080p videos, respectively.
The GoP size in AlphaVC is set to 20 for all testing datasets.

\subsubsection{Camparision.}
Both IPP and LDP configuration of VTM-10.0 and HM-16.20 are used for comparision. 
The IPP only references the previous frame, and each P-frame has the flat QP, which is the same configuration with AlphaVC.
The LDP is the default low-delay P configuration that references multiple previous frames and has dynamic QP for each P-frame.
In addition, state-of-the-art learning-based video compression methods, i.e., 
FVC (CVPR'21)~\cite{hu2021fvc}, DCVC (NIPS'21)~\cite{li2021deep}, B-EPIC (ICCV'21)~\cite{pourreza2021extending}, VLVC (2021)~\cite{feng2021versatile}, TCMVC (2021)~\cite{sheng2021temporal}.
Note that, B-EPIC and VLVC don't belong to IPPP mode, due to the introduction of B-frame.

\begin{figure}[htb]
\centering
\includegraphics[width=\linewidth]{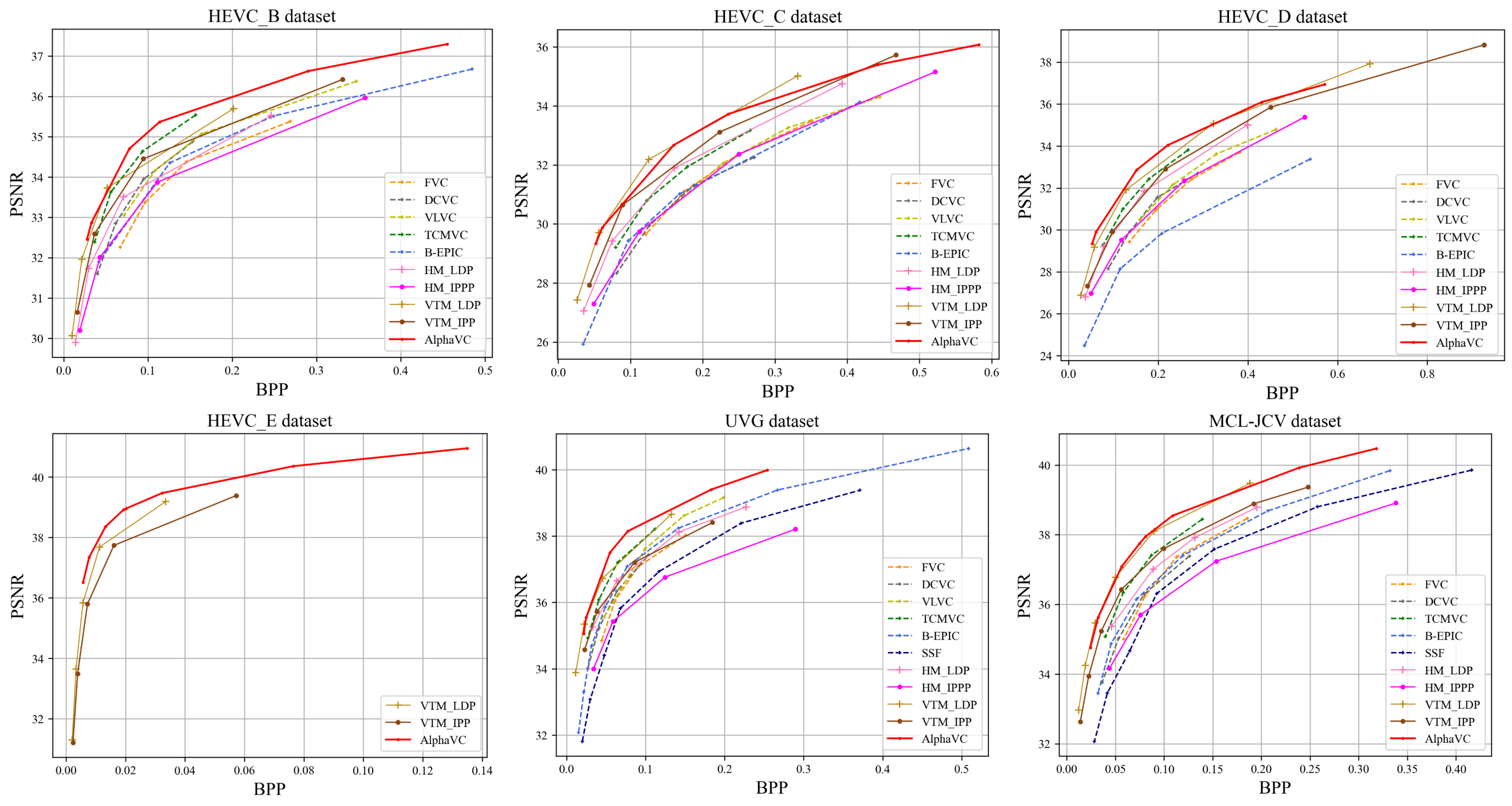}
\caption{PSNR based R-D Curves of traditional codecs and state-of-the-art learning-based codecs on each datasets.
         The red solid line is AlphaVC. Traditional codecs are all represented by solid lines, and other learning-based codecs are represented by dotted lines.}
\label{fig:rdcurve}
\end{figure}

\subsection{Experiment results}
\subsubsection{Performance} 

\begin{figure}[htb]
\centering
\includegraphics[width=\linewidth]{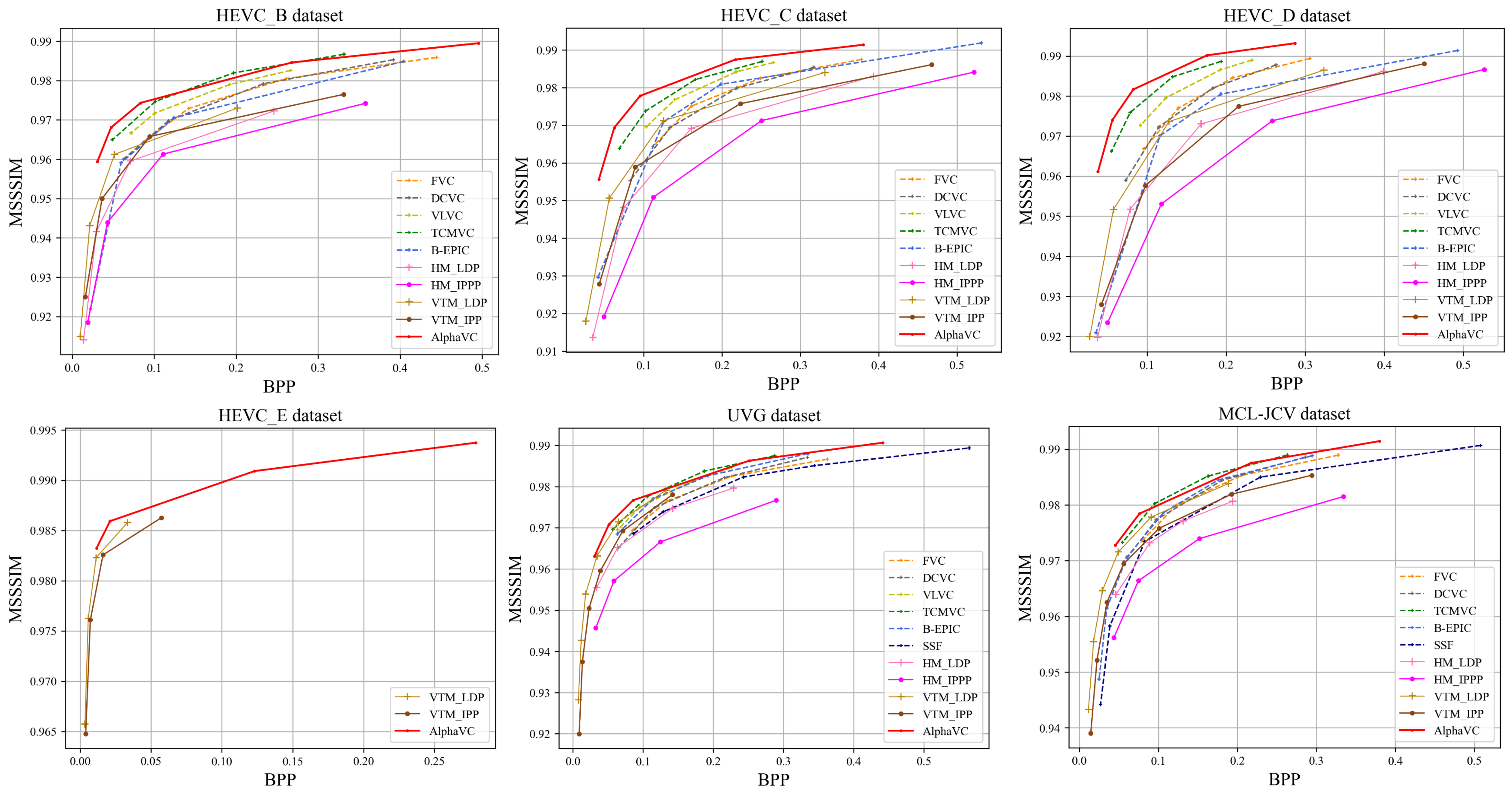}
\caption{MS-SSIM based R-D Curves.}
\label{fig:rdcurve-msssim}
\end{figure}

Fig.~\ref{fig:rdcurve},~\ref{fig:rdcurve-msssim} shows the experimental results on all testing datasets.
It is obvious that AlphaVC achieves the bset performance of all methods. 
In terms of MS-SSIM, AlphaVC significantly outperforms all the other methods over the entire bitrate range and on all the datasets.
In terms of PSNR, AlphaVC significantly outperforms all the learning-based codecs and VTM-IPP, and even outperforms VTM-LDP in most situations.
As mentioned before, VTM-LDP references multiple previous frames and has dynamic QP for each P-frame.
which is not adopted by AlphaVC.

Table~\ref{table:bdrate-psnr} and Table~\ref{table:bdrate-msssim} show the BD-rate savings in PSNR and MS-SSIM that anchored by VTM-IPP.
In terms of PSNR, AlphaVC achieves an average 28.2\% bitrate saving compared to VTM-IPP, outperforming all the reported methods,
including the stronger VTM-LDP (23.5\% bitrate saving). 
In the worst case, AlphaVC also achieves a BD-rate saving of 14.9\% showing a good stability. 
In terms of MS-SSIM, learning-based codecs generally have better performances than traditional codecs,
among with AlphaVC performing the best, by saving an additional 8\% bitrate over the best SOTA TCMVC.

\setlength{\tabcolsep}{4pt}
\begin{table}[h]
\begin{center}
\caption{BD-rate calculated by PSNR with the anchor of VTM-IPP. 
\textcolor{red}{Red} means more bits ($> 3\%$) required. \textcolor{teal}{Green} means fewer bits ($< -3\%$) required.}
\label{table:bdrate-psnr}
\resizebox{\textwidth}{!}{
\begin{tabular}{cccccccccccc}
\hline\noalign{\smallskip}

        & VTM-IPP & VTM-LDP & HM-IPP & HM-LDP & SSF & FVC & DCVC & VLVC & TCMVC & B-EPIC & AlphaVC \\
\noalign{\smallskip}

\hline\noalign{\smallskip}

HEVC\_B &    0    & \textcolor{teal}{-17.9\%} & \textcolor{red}{55.2\%} & \textcolor{red}{24.0\%} & - & \textcolor{red}{75.4\%} 
& \textcolor{red}{43.7\%} & \textcolor{red}{27.1\%} &\textcolor{teal}{-6.92\%}& \textcolor{red}{42.5\%} & \textcolor{teal}{\bf{-22.5\%}}  \\
\noalign{\smallskip}

HEVC\_C &    0    & \textcolor{teal}{\bf{-23.1\%}} & \textcolor{red}{38.6\%} & \textcolor{red}{27.1\%} & - & \textcolor{red}{40.9\%} 
& \textcolor{red}{42.8\%} & \textcolor{red}{40.8\%} & \textcolor{red}{10.2\%} & \textcolor{red}{35.6\%} & \textcolor{teal}{-14.9\%}  \\
\noalign{\smallskip}

HEVC\_D &    0    & \textcolor{teal}{-17.9\%} & \textcolor{red}{35.7\%} & \textcolor{red}{24.9\%} & - & \textcolor{red}{47.9\%} 
& \textcolor{red}{38.6\%} & \textcolor{red}{30.5\%} &\textcolor{teal}{-6.61\%}& \textcolor{red}{117.\%} & \textcolor{teal}{\bf{-29.0\%}}  \\
\noalign{\smallskip}

UVG     &    0    & \textcolor{teal}{-31.9\%} & \textcolor{red}{18.5\%} & \textcolor{black}{1.99\%} & \textcolor{red}{57.7\%} & \textcolor{red}{28.4\%} 
& \textcolor{red}{24.0\%} & 2.15\%                  &\textcolor{teal}{-17.3\%}& \textcolor{red}{3.78\%} & \textcolor{teal}{\bf{-41.7\%}} \\
\noalign{\smallskip}

MCL-JCV     &    0    & \textcolor{teal}{-26.6\%} & \textcolor{red}{26.3\%} & \textcolor{red}{15.2\%} & \textcolor{red}{50.6\%} & \textcolor{red}{29.3\%} 
& \textcolor{red}{43.8\%} & -                  &2.32\%& \textcolor{red}{50.6\%} & \textcolor{teal}{\bf{-32.9\%}} \\
\noalign{\smallskip}

\hline\noalign{\smallskip}

Avg     &    0    & \textcolor{teal}{-23.5\%} & \textcolor{red}{35.6\%} & \textcolor{red}{19.7\%} & \textcolor{red}{54.2\%} & \textcolor{red}{44.4\%} 
& \textcolor{red}{38.6\%} & \textcolor{red}{25.1\%} &\textcolor{teal}{-3.66\%}& \textcolor{red}{49.9\%} & \textcolor{teal}{\bf{-28.2\%}} \\
\noalign{\smallskip}

\hline
\end{tabular}
}
\end{center}
\end{table}
\setlength{\tabcolsep}{1.4pt}

\setlength{\tabcolsep}{4pt}
\begin{table}[h]
\begin{center}
\caption{BD-rate calculated by MS-SSIM with the anchor of VTM-PVC-IPP. 
\textcolor{red}{Red} means more bits ($> 3\%$) required. \textcolor{teal}{Green} means fewer bits ($< -3\%$) required.}
\label{table:bdrate-msssim}
\resizebox{\textwidth}{!}{
\begin{tabular}{cccccccccccc}
\hline\noalign{\smallskip}

        & VTM-IPP & VTM-LDP & HM-IPP & HM-LDP & SSF & FVC & DCVC & VLVC & TCMVC & B-EPIC & AlphaVC \\
\noalign{\smallskip}

\hline\noalign{\smallskip}

HEVC\_B &    0    & \textcolor{teal}{-20.5\%} & \textcolor{red}{54.6\%} & \textcolor{red}{17.4\%} & - & \textcolor{teal}{-21.3\%} 
& \textcolor{teal}{-16.0\%} & \textcolor{teal}{-42.5\%} &\textcolor{teal}{-53.5\%}& \textcolor{teal}{-7.1\%} & \textcolor{teal}{\bf{-61.6\%}}  \\
\noalign{\smallskip}

HEVC\_C &    0    & \textcolor{teal}{-20.7\%} & \textcolor{red}{53.6\%} & \textcolor{red}{12.8\%} & - & \textcolor{teal}{-22.2\%} 
& \textcolor{teal}{-12.8\%} & \textcolor{teal}{-41.6\%} & \textcolor{teal}{-47.6\%} & \textcolor{red}{-15.4\%} & \textcolor{teal}{\bf{-58.9\%}}  \\
\noalign{\smallskip}

HEVC\_D &    0    & \textcolor{teal}{-27.2\%} & \textcolor{red}{39.3\%} & {-1.5\%} & - & \textcolor{teal}{-34.7\%} 
& \textcolor{teal}{-33.0\%} & \textcolor{teal}{-49.6\%} &\textcolor{teal}{-60.7\%}& \textcolor{teal}{-21.5\%} & \textcolor{teal}{\bf{-67.2\%}}  \\
\noalign{\smallskip}

UVG     &    0    & \textcolor{teal}{-26.7\%} & \textcolor{red}{56.3\%} & \textcolor{red}{20.2\%} & \textcolor{red}{33.9\%} & \textcolor{red}{11.5\%} 
& \textcolor{red}{10.9\%} & \textcolor{teal}{-12.9\%} &\textcolor{teal}{-22.0\%}& -1.63\%       & \textcolor{teal}{\bf{-32.9\%}} \\
\noalign{\smallskip}

MCL-JCV  &    0    & \textcolor{teal}{-26.0\%} & \textcolor{red}{49.6\%} & \textcolor{red}{14.5\%} & \textcolor{teal}{-4.5\%}& \textcolor{teal}{-18.8\%} 
&\textcolor{teal}{-17.9\%}& -                       &\textcolor{teal}{-38.8\%}&\textcolor{teal}{-19.9\%}& \textcolor{teal}{\bf{-40.5\%}} \\
\noalign{\smallskip}

\hline\noalign{\smallskip}

Avg     &    0    & \textcolor{teal}{-24.2\%} & \textcolor{red}{49.9\%} & \textcolor{red}{11.5\%} & \textcolor{red}{14.7\%} & \textcolor{teal}{-17.1\%} 
& \textcolor{teal}{-13.7\%} & \textcolor{teal}{-36.6\%} &\textcolor{teal}{-44.5\%}& \textcolor{teal}{-13.1\%} & \textcolor{teal}{\bf{-52.2\%}} \\
\noalign{\smallskip}

\hline
\end{tabular}
}
\end{center}
\end{table}
\setlength{\tabcolsep}{1.4pt}

\subsubsection{Complexity} 
The MAC(Multiply Accumulate) of the P-frame at the decoding side is about 1.13M/pixel, and the cI-frame is about 0.98M/pixel. 
We use arithmetic coding for the complete entropy encoding and decoding process, and 1080p videos to evaluate the runtime. 
The runtime of the encoding side includes model inference, data transmission from GPU to CPU and entropy encoding, 
and the runtime of the decoding side includes entropy decoding, data transmission and model inference. 
The comparison results are shown in Table~\ref{table:runtime}, 
in which running platform of AlphaVC is Intel(R) Xeon(R) Gold 6278C CPU and NVIDIA V100 GPU. 
The encoding and decoding times of AlphaVC on a 1080p frame average about 715ms and 379ms. 
The encoding time is about 1000x faster than VTM, and the decoding time is similar to VTM (1.69x). 
Even though AlphaVC uses more parameters than TCMVC, it is still faster. 
The main reason is the proposed probability-based skip entropy technique, 
which significantly reduces the running time on CPU. 
In addition, we can find that the cI-frame is slower than P-frame although the cI-frame has less complexity. 
This is also because the bit-rate in the cI-frame is higher, and the number of skipping elements in the cI-frame is fewer.

\setlength{\tabcolsep}{4pt}
\begin{table}[h]
\begin{center}
\caption{
  Complexity on 1080p video. We compare our AlphaVC including cI-Frame and p-Frame with traditional codecs and TCMVC. 
  The time ratio is calculated with the anchor of VTM.
}
\label{table:runtime}
\resizebox{0.8\textwidth}{!}{
\begin{tabular}{cccccc}
\hline\noalign{\smallskip}

Method & Params. & Enc-T (s) & Dec-T (s) & Enc-T ratio & Dec-T ratio \\
\noalign{\smallskip}

\hline\noalign{\smallskip}

VTM-10.0-IPP   &   -     & 661.9 & 0.224 & 1.0000 & 1.0000 \\
\noalign{\smallskip}
HM-16.40-IPP   &   -     & 26.47 & 0.140 & 0.0400 & 0.6250 \\
\noalign{\smallskip}
TCMVC      &   10.7M & 0.827 & 0.472 & 0.0012 & 2.1071 \\
\noalign{\smallskip}
\hline\noalign{\smallskip}
AlphaVC    &   63.7M & 0.715 & 0.379 & 0.0011 & 1.6920 \\
\noalign{\smallskip}
AlphaVC-cI &   29.9M & 0.733 & 0.580 & 0.0011 & 2.5893 \\
\noalign{\smallskip}
AlphaVC-P  &   33.8M & 0.685 & 0.365 & 0.0010 & 1.6295 \\
\noalign{\smallskip}

\hline
\end{tabular}
}
\end{center}
\end{table}
\setlength{\tabcolsep}{1.4pt}

\subsection{Ablation Study and Analysis}
\subsubsection{Frame Analysis}
\begin{figure}[htb]
\centering
\includegraphics[width=\linewidth]{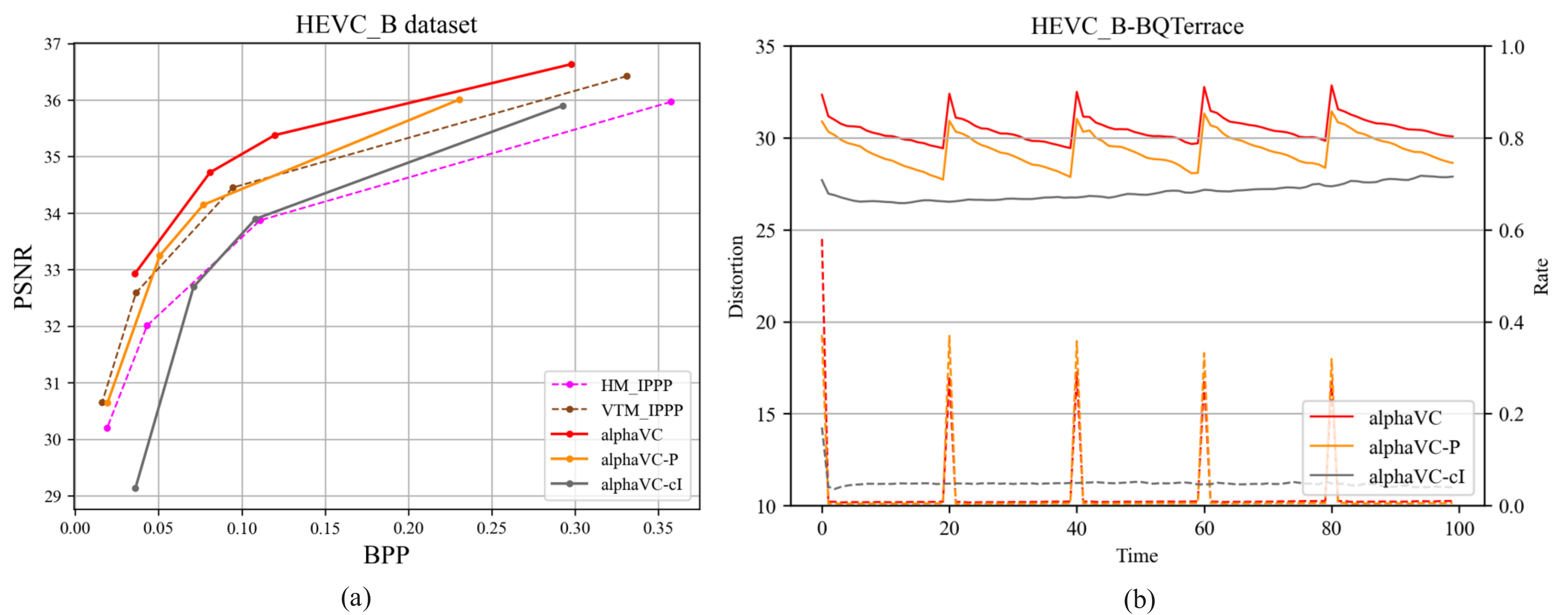}
\caption{
  Comparison with each type of frame in AlphaVC. 
  AlphaVC-P only include P-frame and I-frame, the GoP size is 20 samed as AlphaVC.
  AlphaVC-cI only include cI-frame and I-frame, only the first frame uses the I-frame.
  (a): R-D performance of AlphaVC, AlphaVC-P and AlphaVC-cI under PSNR on HEVC class B dataset. 
  (b): Example of performance comparison for each type of frame, the tested sequence is BQTerrace in class B. 
  The solid line indicates the curve of distortion, the dashed line indicates the curve of rate.
}
\label{fig:curve-frames}
\end{figure}

We use three types of frame in AlphaVC:I-frame, cI-frame and P-frame.
To justify this approach and evaluate each type of frame, 
we train two additional models AlphaVC-P and AlphaVC-cI. 
AlphaVC-P only includes I-frame and P-frame, and the GoP size is the same with AlphaVC in the test phase.
AlphaVC-cI only includes I-frame and cI-frame, and there is no group in AlphaVC-cI, I-frame is only used in the first frame and all subsequent frames are cI-frames.
The R-D performance is shown in Fig.~\ref{fig:curve-frames}(a), AlphaVC-P achieves comparable performance with VTM\_IPP, and AlphaVC-cI only achieves comparable performance with HM\_IPP. 
The reason may be that cI-frame utilizes reference frames in a more implicityly way: as the condition of entropy.
The reason is that, although the cI-frame is not good enough, it is stable and has no accumulated error as shown in Fig.~\ref{fig:curve-frames}(b).
By combining these two types of frame, AlphaVC achieves better R-D performance for the following two reasons:
\begin{enumerate}
  \item The accumulated error of P-frame in AlphaVC is smaller than the P-frame in AlphaVC-P. (see in Fig.~\ref{fig:curve-frames}(b)).
  \item The performance of cI-frame is much better than I-frame (see in Fig.~\ref{fig:curve-frames}, similar distortion with smaller rate).
\end{enumerate}

\subsubsection{Effectiveness of Different Components.}

\begin{figure}[t]
\centering
\includegraphics[width=\linewidth]{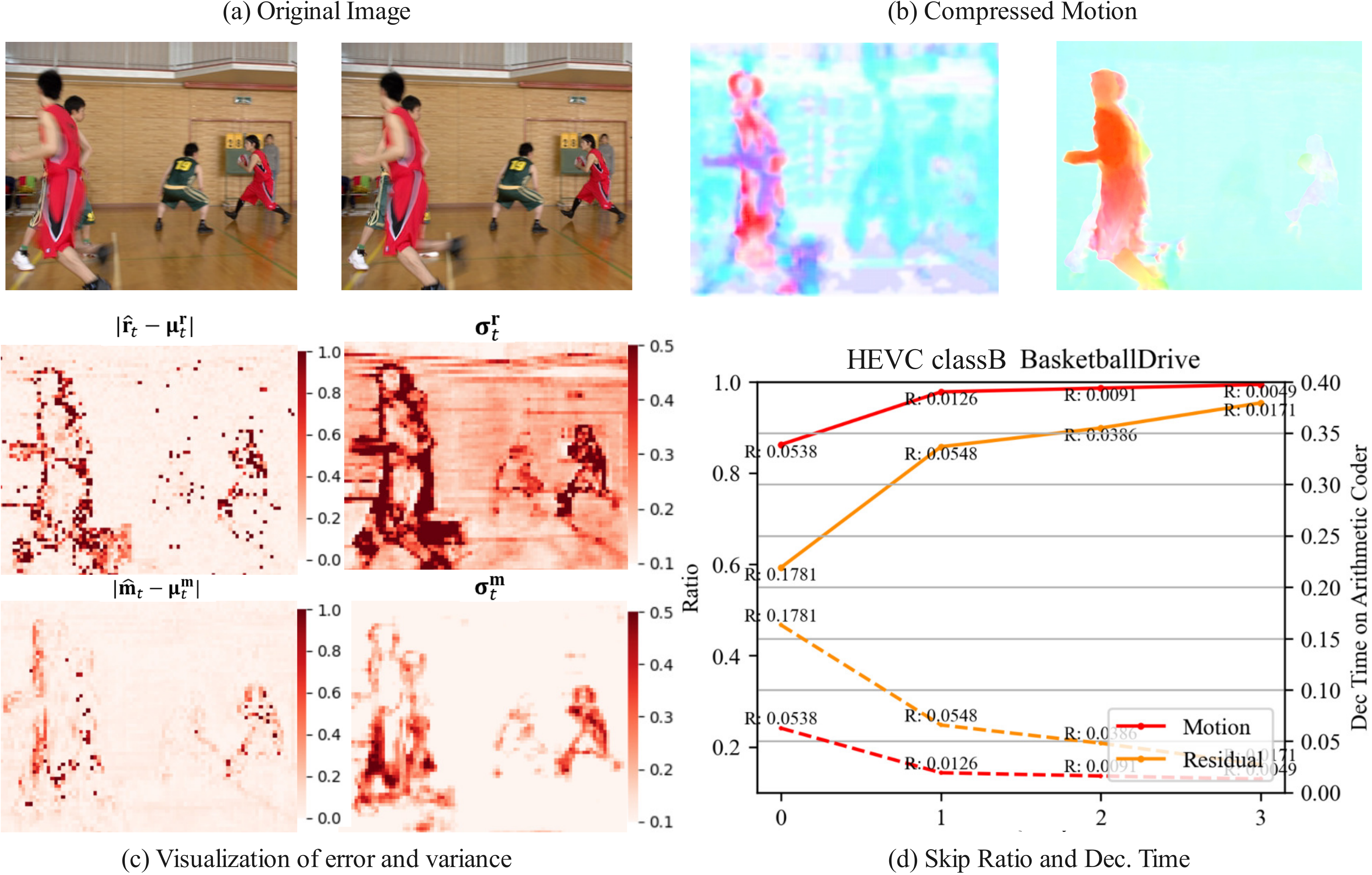}
\caption{
  Analysis of methods.
  (a): Two adjacent original frames of HEVC classB BasketballDrive.
  (b): Left/Right: The compressed motion wo/w our motion prediction module.
  (c): Visualization of variance of gaussian distortion $\bf{\sigma}$ and error after replacement.
  (d): Example result of the average skip ratio and arithmetic decoding time at 4 different bit rates,
  the ratio is calculated by skipped elements / total elements. 
  The motion and residual latents are shown in the red and yellow curve, respectively.
  The solid and dotted curves represent ratio and time, respectively.
  The number on curves indicates bit-rate(BPP).
}
\label{fig:skip-prob}
\end{figure}

We demonstrate the effectiveness of our proposed components with AlphaVC-P as the anchor.
We gradually remove the P2F MP, Skip in $\bf{\hat{m}}$ and Skip in $\bf{\hat{r}}$ from AlphaVC-P. 
Note that, without P2F MP, the current feature and reference feature will be fed to the motion compression module directly.
The BD-Rate savings against AlphaVC-P are presented in Table~\ref{table:ablation}(b). 
Moreover, a more intuitive analysis for the proposed methods is shown in Fig.~\ref{fig:skip-prob}.

As shown in  Table~\ref{table:ablation}(b), P2F MP brings 10.4\% BD-rate saving.
From Fig.~\ref{fig:skip-prob}(b), we can see that the compressed motion with P2F MP is more accurate and with smaller entropy.

\setlength{\tabcolsep}{4pt}
\begin{table}[htb]
  \caption{
    Effectiveness of our different components. 
    The BD-rate values are computed under PSNR on HEVC class B dataset.
  }
  \label{table:ablation}
  \begin{subtable}{0.45\linewidth}
    \caption{}
    \centering
    \resizebox{0.8\textwidth}{!}{
    \begin{tabular}{lccc}
    \hline\noalign{\smallskip}
    I-frame  & \checkmark & \checkmark & \checkmark \\
    P-frame  & \checkmark & \checkmark & \\
    cI-frame & \checkmark &  & \checkmark \\
    \hline\noalign{\smallskip}
    BD-Rate & 0\% & 21.4\% & 92.7\%  \\
    \hline\noalign{\smallskip}
    \end{tabular}
    }
  \end{subtable}
  \begin{subtable}{0.55\linewidth}
    \caption{}
    \centering
    \resizebox{0.85\textwidth}{!}{
    \begin{tabular}{lcccc}
    \hline\noalign{\smallskip}
    P2F MP & \checkmark &            & & \\
    Skip in M.  & \checkmark & \checkmark & & \\
    Skip in R.& \checkmark & \checkmark & \checkmark & \\
    \hline\noalign{\smallskip}
    BD-Rate & 0\% & 10.4\% & 18.6\% & 37.5\% \\
    \hline\noalign{\smallskip}
    \end{tabular}
    }
  \end{subtable}
\end{table}
\setlength{\tabcolsep}{1.4pt}

To analyze Skip, 
we first explore the relationship between the replacement error, and the variance of Gaussian distribution as shown in Fig.~\ref{fig:skip-prob}(c).
Notice that the replacement error is highly correlated with variance, and elements with smaller variance have small errors.
Therefore, skipping the entropy coding of these elements will not cause any loss, and may even improve performance. 
Due to the smoothness of motion information, 
the Skip ratio of motion latents is as high as 90\% at each quality level as shown in Fig.~\ref{fig:skip-prob}(d),
The Skip ratio of residual latents gradually increases (60\% -- 90\%) with the decrease of quality.
With the number of skipped elements increases, we can clearly see  in Fig.~\ref{fig:skip-prob}(d) that the runtime of entropy coding on CPU is greatly reduced.
In addition, as shown in Table~\ref{table:ablation}(b), the probability-based skip entropy method can also improve performance obviously.

\section{Conclusion}
This paper proposed a high-performance and efficient learned video compression approach named AlphaVC.
Specifically, we designed a new coding mode including three types of frame: I-frame, P-frame, and cI-frame,
to reduce the bit rate of I-frame and mitigate the accumulative error.
We then proposed two efficient techniques: 
P2F MP for improving the accuracy of inter-frame prediction at the encoder side,
and Skip for reducing entropy and speeding up runtime.
Experimental results show that AlphaVC outperforms H.266/VVC in terms of PSNR by 28\% under the same configuration, 
meanwhile AlphaVC has the comparable decoding time compared with VTM.
To the best of our knowledge, AlphaVC is the first learned video compression scheme achieving such a milestone result that 
outperforms VTM-IPP over the entire bitrate range and on all common test datasets.

We believe that our proposed AlphaVC provides some novel and useful techniques that
can help researcheres to further develop the next generation video codecs with more powerful compression.

\clearpage
%
%
\bibliographystyle{splncs04}
\bibliography{alphaVC}
\end{document}